# A Double-Deep Spatio-Angular Learning Framework for Light Field based Face Recognition

Alireza Sepas-Moghaddam, *IEEE, Student Member,* Mohammad A. Haque, *IEEE, Member,*
Paulo Lobato Correia, *IEEE, Senior Member,* Kamal Nasrollahi, *IEEE, Member,*
Thomas B. Moeslund, Fernando Pereira, *IEEE, Fellow*

*Abstract*— Face recognition has attracted increasing attention due to its wide range of applications, but it is still challenging when facing large variations in the biometric data characteristics. Lenslet light field cameras have recently come into prominence to capture rich spatio-angular information, thus offering new possibilities for advanced biometric recognition systems. This paper proposes a double-deep spatio-angular learning framework for light field based face recognition, which is able to model both the intra-view/spatial and inter-view/angular information using two deep networks in sequence. This is a novel recognition framework that has never been proposed in the literature for face recognition or any other visual recognition task. The proposed double-deep learning framework includes a long short-term memory (LSTM) recurrent network whose inputs are VGG-Face descriptions, computed using a VGG- 16 convolutional neural network (CNN). The VGG-Face spatial descriptions are extracted from a selected set of 2D sub-aperture (SA) images rendered from the light field image, corresponding to different observation angles. A sequence of the VGG-Face spatial descriptions is then be analysed by the LSTM network. A comprehensive set of experiments has been conducted using the IST-EURECOM light field face database, addressing varied and challenging recognition tasks. Results show that the proposed framework achieves superior face recognition performance when compared to the state-of-the-art.

## I. INTRODUCTION

Face recognition systems have been successfully used in various application areas, ranging from forensics and surveillance to commerce and entertainment [1] [2] [3]. With the development of deep learning solutions and the increase in computational power, rapid advances in a variety of visual recognition tasks, including face recognition, have been observed in recent years. Nowadays, the state-of-the-art on face recognition is dominated by deep neural networks [4] [5]. However, even with the emergence of this type of sophisticated networks, certain conditions may still not yet allow achieving accurate enough face recognition, notably because the acquisition process may introduce challenging variations in the biometric data, especially in less constrained scenarios where it is expected to find significant variations in terms of emotions, poses, illumination, occlusions, and aging, among others [6].

The emergence of new imaging sensors such as depth, near infra-red (NIR), thermal, and lenslet light field cameras, is opening new frontiers for face recognition systems [1]. Naturally, the richer scene representations captured by these emerging imaging sensors may contribute to boost the face recognition performance. Lenslet light field cameras have recently come into prominence as they are able to capture the intensity of the light rays coming from multiple directions [7] [8], thus allowing to take benefit from the richer spatio-angular information available. Light field cameras have been successfully applied, not only to face recognition [9] [10] [11] [12] [13] [14] [15] but also to face presentation attack detection (PAD) [16] [17] [18] [19] [20] [21].

This paper proposes a double-deep spatio-angular learning framework where two deep learning networks are used in sequence, in an end-to-end face recognition architecture, exploiting the intra-view and inter-view information available in the set of views associated to a lenslet light field image, hereafter referred only as a 'light field image' for simplification. The proposed end-to-end recognition architecture first learns deep intra-view/spatial descriptions using the VGG-Face descriptor [22]. The VGG-Face descriptions are computed using a VGG-16 CNN architecture [23] as this is one of the most efficient and commonly used CNN models for deep face description. The framework then includes an LSTM recurrent neural network [24] that is trained to exploit the available inter-view/angular information conveyed by the VGG-Face descriptions for a selection of 2D viewpoint images rendered from the light field. In practice, the combination of the VGG-Face and LSTM networks is able to learn powerful deep representations from the spatial and angular information present in the lenslet light field images.

This is a novel framework for face recognition or any other visual recognition task, acknowledging that the additional angular information brings complementary information contributing to boost the recognition performance. The proposed recognition solution is generic and powerful enough to address rather extreme scenarios, e.g. occlusions, and aging [25] [26] [27], while still achieving a very good recognition performance. This is largely due to the proposed double-deep learning framework, which is able to learn a more powerful and complete model, taking into account the richer facial information captured from the different observation viewpoints available in light field images.

The proposed face recognition solution takes as input a raw light field face image (captured with a lenslet light field camera) to create a multi-view array, composed by a set of 2D sub-aperture (SA) images, each SA corresponding to a slightly different viewpoint, i.e., observing the face from a different angle. A selected set of representative SA images are organized as a sequence, and a VGG-Face description is created for each SA image. Then, the sequence of extracted VGG-Face deep descriptions, each corresponding to a different angular position, is the input to an LSTM network; finally, a softmax layer is used for classification.

The double-deep spatio-angular representation has been evaluated using the IST-EURECOM Light Field Face Database (LFFD) [28], which includes different facial variations such as emotions, poses, illuminations and occlusions. As the original paper presenting the IST-EURECOM LFFD [28] does not propose any test protocol for performance assessment, this paper proposes also two novel, practically meaningful, test protocols to conduct face recognition experiments on the LFFD database. The proposed test protocols are essential to assess the overall performance of the proposed recognition solution as well as to study its sensitivity to the available training data, both in terms of number of training samples and facial variations. A comprehensive validation experiment is performed to optimally configure the proposed end-to-end recognition architecture in terms of accuracy, network complexity, convergence speed, and learning and testing times. Several experiments have been conducted to assess the recognition performance of the proposed framework in comparison with 17 state-of-the-art benchmarking solutions. Results show that the proposed double-deep learning recognition framework offers a powerful solution, providing significant improvements in face recognition performance, even when considering the operation in challenging test conditions.

The rest of the paper is organized as follows: Section II briefly reviews the basic concepts of light field imaging, existing light field based face recognition solutions, and current deep CNN architectures for face recognition. The proposed double-deep spatio-angular learning framework for face recognition is presented in Section III. Section IV describes the experimental setup and assessment methodology. The performance evaluation and associated analysis are presented in Section V. Finally, Section VI concludes the paper.

## II. BACKGROUND

This section briefly reviews the basic concepts of light field imaging, existing light field based face recognition solutions, and CNN based face recognition solutions.

### A. Light Field Imaging Basics

The proposed face recognition solution takes benefit from a new type of visual sensor, the so-called *lenslet light field cameras*. The captured light field image allows the development of face recognition solutions exploiting additional visual information, and achieving better recognition performance.

The so-called *plenoptic function* $P(x,y,z,t,\lambda,\theta,\varphi)$ was proposed in 1991 to model the information carried by the light rays at every point in the 3D space $(x,y,z)$, for every possible direction $(\theta, \varphi)$, over any wavelength $(\lambda)$, and at any time $(t)$ [29]. The so-called *static 4D light field* [30], $L(x,y,u,v)$, also known as *lumigraph* [31], was proposed in 1996, by adopting several simplifications on the plenoptic function and may be described by the intersection points of the light rays with two parallel planes [32].

There are currently two main practical setups for capturing light fields: i) a high density array of regular cameras, such as the Stanford multi-camera arrays [33] and the JPEG Pleno high density camera array (HDCA) [34]; and ii) a lenslet light field camera, using an array of micro-lenses placed in front of an image sensor to capture the light information for different incident angles [35].

Regarding lenslet light field cameras, there are currently two main types, the so-called *plenoptic 1.0* and *plenoptic 2.0* cameras. The most common type is the plenoptic 1.0 camera, the one considered in this paper, and in the following simply called *light field camera*. In this type of camera, the main lens is focused on the micro-lens plane while the micro-lenses are focused at infinity. Unlike a conventional 2D camera that captures an image by integrating the intensities of all rays (from all directions) impinging each sensor element, in the light field camera each pixel collects the light of a single ray (or a thin bundle of rays) from a given angular direction, converging on a specific micro-lens in the array.

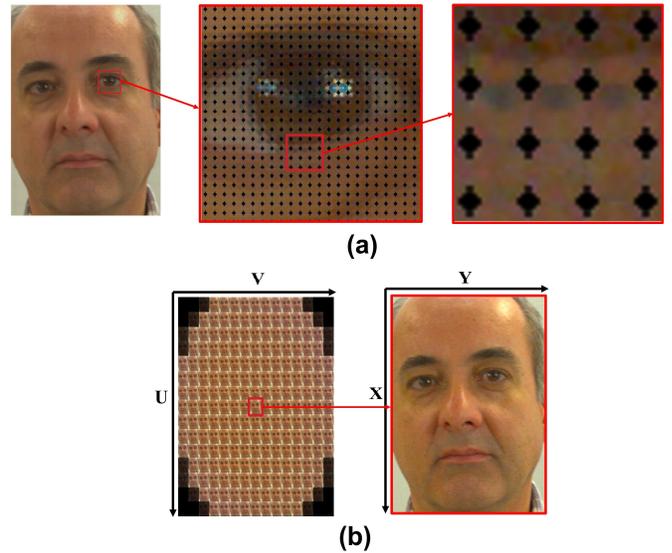

**Figure 1:** Light field representation: (a) sample micro-images, after colour demosaicing; (b) corresponding multi-view SA array.

In a lenslet light field camera, each micro-lens acquires a micro-image with a Bayer pattern filter; thus, a demosaicing operation is needed to convert micro-images into the RGB colour space; Figure 1 (a) shows sample micro-images, after colour demosaicing. The demosaiced light field image (made of micro-images) can then be rendered to form a multi-view SA array with size $U \times V \times X \times Y \times 3$, where $U \times V$ corresponds to the number of views, $X \times Y$ corresponds to the spatial resolution of each resulting 2D SA image, and the '3' corresponds to the R, G and B color components, as illustrated in Figure 1 (b); the black SA images in the corners are due to the so-called *vignetting effect*.

The IST-EURECOM LFFD images used in this research work were captured with a Lytro Illum camera [35] [36], where the corresponding multi-view SA array includes 15×15 SA images, each with a spatial resolution of 625×434 pixels. Each 2D SA image corresponds to a slightly different viewpoint of the visual scene, thus 'seeing' the visual scene from a slightly different observation angle. The resulting angular information is a distinctive characteristic of this new type of visual sensor.

Since one light field image allows obtaining multiple 2D SA images, two types of information can be learned from it: i) the spatial, textural, intra-view information within each view; and ii) the angular, inter-view information associated to the different angles of observation of the multiple views.

It should be noted that the term 'angular' does not mean that angle values are processed but rather that angular dependent information/intensities are processed. The same happens when referring to 'spatial' information as no position coordinate values are processed but rather position dependent information/intensities are processed.

A major novelty of this paper is precisely to exploit/learn/describe both the spatial information (within each view) and the angular information (across views), in a way that is not yet available in the literature, and which is only possible because a lenslet light field camera is used.

### B. Prior Light Field based Face Recognition Solutions

As this paper focuses on light field based face recognition, this section reviews the face recognition solutions in the literature already exploiting light field sensors.

Several face recognition solutions exploiting the richer light field imaging information have recently been proposed. Table I summarizes the main characteristics of prior light field based face recognition solutions along with the used databases, sorted according to their publication date. The genesis of these solutions is associated to three distinctive light field capabilities, i.e., *a posteriori* refocusing, disparity exploitation and depth exploitation. The solutions summarized in Table I are briefly reviewed in the following, grouped according to the exploited light field capability.

- **Solutions relying on *a posteriori* refocusing**

Based on the intra-view, angular information available in a light field image, *a posteriori* refocusing to a given selected plane can be supported. This is useful to improve the quality of a previously out-of-focus *region of interest* for the subsequent recognition of either a single face or multiple faces, positioned at different distances.

Raghavendra *et al.* [9] proposed a wavelet energy based approach to select the best focused face image from a set of refocused images rendered from a light field image. Later on, the same authors proposed a resolution enhancement scheme based on the discrete wavelet transform to capture high frequency components from different focused 2D images, rendered from a single light field image [10]. Raghavendra *et al.* [11] also investigated the identification of multiple faces at different distances by exploiting an all-in-focus image created from a light field image. The same group has also proposed a face recognition solution relying on rendering a light field image at different focus planes in two different ways: i) selecting the best focused image; and ii) combining focused images to create a super-resolved image. Both approaches have been considered for face recognition [12].

- **Solutions relying on disparity exploitation**

A light field image can be structured as a set of 2D SA images, each corresponding to a specific scene viewpoint. This representation includes information about disparity that can be exploited to improve the face recognition performance.

Sepas-Moghaddam *et al.* [13] proposed a Light Field Local Binary Patterns (LFLBP) descriptor with two main components: (i) a conventional Spatial Local Binary Pattern (SLBP), corresponding to the local binary patterns for the central SA image; and (ii) a novel Light Field Angular Local Binary Pattern (LFALBP), able to represent the variations associated to the different directions of light captured in light field images. The combination of these complementary descriptors improved the face recognition accuracy. In [14], the same authors proposed a new light field based ear recognition solution that can also be used for faces or other biometric traits. It is based on the fusion of a non-light field based descriptor, the Histogram of Oriented Gradients (HOG), with a light field based descriptor, the Histogram of Disparity Gradients (HDG). By exploiting texture and disparity, the overall recognition performance was improved.

**Table 1**: Overview of prior light field based face recognition solutions. *The abbreviations used in this table*: BSIF, Binarized Statistical Image Features; CSLBP, Centre-Symmetric Local Binary Patterns; HDG, Histogram of Disparity Gradient; HOG, Histogram of Oriented Gradient; LBP, Local Binary Pattern; LF, Light Field; LFLBP, Light Field Local Binary Pattern; LFFD, Light Field Face DataBase; LLFEDB, Lenslet Light Field Ear DataBase; LG, Log Gabor; M-V, Multi-View; NNC, Nearest Neighbor Classifier; SRC, Sparse Representation Classifier; SVM, Support Vector Machine; WE, Wavelet Energy.

| Ref. | Year | Feature Extraction Method | Classifier | Light Field Capability | Format | Database |
|---|---|---|---|---|---|---|
| [9] | 2013 | WE | NNC | Depth computation | 2D rendered from LF | Private |
| [10] | 2013 | LBP; LG filter | SRC | *A posteriori* refocusing | 2D rendered from LF | Private |
| [11] | 2013 | LBP | SCR | *A posteriori* refocusing | 2D rendered from LF | Private |
| [12] | 2016 | HOG; LBP; CSLBP; BSIF | SRC | *A posteriori* refocusing | 2D rendered from LF | Public: GUC-LiFFID |
| [37] | 2016 | HOG | SVM | Depth computation | M-V SA array | Private |
| [13] | 2017 | LFLBP | NNC | Disparity exploitation | M-V S-A array | Public: LFFD |
| [14] | 2018 | HOG+HDG | NNC | Disparity exploitation | M-V SA array | Public: LLFEDB |
| [15] | 2018 | VGG-Face descriptor | SVM | Disparity exploitation; Depth computation | M-V SA array | Public: LFFD |

- **Solutions relying on depth exploitation**

Considering the available disparity information and the camera intrinsic parameters, it is possible to extract a depth map from the light field image. A face depth map provides geometric information about the position and shape of facial components, which can be exploited for face recognition.

Shen *et al.* [37] extracted a depth map from a light field image and applied a histogram of oriented gradients descriptor for extracting discriminative features, which were then fed into a linear SVM classifier to perform the face recognition task.

- **Solutions relying on more than one light field capability**

Sepas-Moghaddam *et al.* [15] proposed the first CNN based method for light field face recognition relying on both disparity and depth map exploitation, extracted using independent approaches. The proposed solution independently extracts disparity and depth maps from the multi-view SA images to be fed into a VGG-16 CNN architecture for fine-tuning the model, which has been pre-trained for texture. The VGG-Face descriptor is used to extract features from the 2D central SA image and from the disparity and depth maps, as they may express some visually complementary information for face recognition, notably if independently extracted. The features extracted for each data type are concatenated and a SVM classifier is applied to the fused deep representation for recognition.

*C. Constraint-aware Solutions for Extreme Face Recognition*

Although face recognition is a widely researched area, with solutions ranging from the classical eigenfaces [38] to state-of-the-art deep learning based recognition approaches [3], this task is still very challenging due to the large variations in pose, illumination, occlusions and aging in real-world scenarios [39] [40]. In fact, it has been revealed that the appearance changes caused by extreme facial and environmental variations significantly surpass the intrinsic differences between individuals [1]. Hence, some constraint-aware face recognition solutions have been proposed to address rather extreme scenarios.

To bridge the cross-pose gap for face recognition, several pose-invariant face recognition solutions have been proposed, notably extracting pose-robust features using regression-based solutions [41], deep CNNs [42], and face frontalization deep networks [43]. Another important challenge for practical face recognition solutions is their robustness to uncontrolled lighting conditions [44]; to address this problem, a number of illumination-invariant face recognition solutions have been adopted, notably shadow suppression illumination normalization [45], autoencoder neural networks [46], and generative adversarial networks [47]. Occlusion is another major challenge since facial images are often occluded by facial accessories, objects in front of the face, and shadows, thus degrading facial features discrimination and the consequent face detection and recognition performance [48]. Examples of occlusion-invariant recognition solutions include dynamic image-to-class warping [49], and matrix regression [50]. Another problem is related to facial aging [51]. Some age-invariant face recognition solutions include dictionary learning [25] [27] and latent factor guided CNN [26]. Finally, it is worth noting that the face recognition solution proposed in this paper is a generic solution, learning a powerful and complete spatio-angular model, rather than only addressing face recognition under extreme variations; however, the performance results show that it can work very well for rather extreme cases.

*D. Current Deep CNN Architectures for Face Recognition*

In recent years, deep learning architectures have been increasingly adopted for face recognition tasks. Deep CNN architectures take raw data as their input, and extract features using convolutional filters in multiple levels of abstraction. However, optimizing tens of millions of weights to learn deep learning weights needs a huge amount of labeled samples along with powerful computational resources. Deep learning models, extracted using CNN architectures, are optimized based on previously labelled data, and then used for feature extraction and classification. Nowadays, the most efficient and commonly used CNN architectures for face recognition are AlexNet [52], Lightened CNN [53], SqueezeNet [54], GoogLeNet [55], and VGG-Very-Deep-16 [23].

In [56], Ghazi *et al.* presented a comprehensive evaluation of deep learning models for face recognition, computed using the above mentioned CNN architectures, under various facial variations. Additionally, the impacts of different covariates, such as compression artefacts, occlusions, noise, and color information, on the recognition performance of the above mentioned architectures have been studied by Grm *et. al.* [57]. The results have shown that the VGG-Face descriptor [22], computed based on a VGG-16 CNN architecture, achieves superior recognition performance under various facial variations, and is more robust to different covariates, when compared to alternatives. This justifies the usage of the VGG-Face descriptor to learn deep facial texture representations in this work and thus its inclusion in the proposed double-deep learning framework.

III. PROPOSED DOUBLE-DEEP LEARNING FRAMEWORK FOR LIGHT FIELD BASED FACE RECOGNITION

This section presents the proposed face recognition framework, exploiting the spatio-angular information available in light field images by using two deep learning networks.

*A. Double-Deep Spatio-Angular Learning Framework*

The double-deep spatio-angular learning framework for light field based face recognition proposed in this paper is based on a novel combination of a deep VGG-Face descriptor with a deep LSTM recurrent neural network, thus justifying the 'double-deep' term used in the name of the proposed solution. While the combination of VGG and LSTM has recently been used to learn spatio-temporal (video) models for visual classification and description tasks, including action recognition [58], facial expression classification [59], or image captioning and video description [60], this combination has never been proposed to exploit the inter-view, angular information available in a light field image captured at a single temporal instant, as proposed in this paper. To this end, the proposed double-deep framework includes novel modules for SA image selection and scanning, targeting the creation of a sequence composed by VGG-Face descriptions of SA images captured from different angles. The VGG-Face descriptions are extracted from each of the selected SA images, and the sequence of descriptions is provided as input to the LSTM network, which then learns the inter-view,

5angular model. Hence, the proposed double-deep CNN-LSTM combination can be very powerful to jointly exploit the spatio-angular information available in light field images, targeting boosting the face recognition performance.

### B. Architecture and Walkthrough

The proposed end-to-end learning architecture is represented in Figure 2 and includes the main modules described in the following.

1. **Pre-processing:** The Light Field Toolbox v0.4 software [61] has been used to create the multi-view SA array, $L(u,v,x,y)$, from the light field raw (LFR) input image, as discussed in Section II.A. Then, the face region is cropped within each SA image in the multi-view array, based on the landmarks provided in the used database. Finally, the cropped SA images are resized to 224×224 pixels as this is the input size expected by the VGG-Face descriptor.
2. **SA image selection and scanning:** This module successively scans a selected sub-set of the SA images into a SA image sequence, as discussed in Section III-C.
3. **VGG-Face spatial extraction:** Each selected SA image, corresponding to a slightly different viewpoint angle, is fed into a *pre-trained* VGG-16 network to extract a spatial description containing 4096 elements, as discussed in Section III-D. Since a pre-trained model has been used, no additional VGG-Face training/learning has been performed for the specific purposes of this paper, and so there is no need to set any training parameter. The spatial descriptions are extracted from the Fully-Connected 6 (FC6) layer of VGG-Face network, which contains 4096 elements, for each of the VGG-Face networks considered, one per 2D SA image.
4. **LSTM angular extraction:** The extracted intra-view, spatial deep descriptions are input to an LSTM network with peephole connections, to learn the model expressing the inter-view, angular dependencies across the selected SA viewpoints and then extracting double-deep spatio-angular descriptions for classification, as discussed in Sections III-E and III-F.
5. **Softmax classification**: The set of outputs from the LSTM gates, corresponding to the view-point changes observed so far, is given as input to a softmax classifier. Then, the average of the classification probabilities across the SA images, selected from the light field image, is used to predict the most probable label and thus the final recognition output, as discussed in Section III-G.

### C. SA Image Selection and Scanning

The multi-view SA array contains 15×15 2D SA images, each corresponding to a slightly different viewpoint angle. A representative subset of SA images is selected for processing by the VGG-Face descriptor, and then scanned as a sequence so that their inter-view, angular model can be learned by the LSTM network. Different methods can be considered to select and scan the set of SA images, notably varying their number, position and scanning order. It is worth noticing that since the Lytro Illum camera microlens shape is hexagonal, there is the so-called *vignetting effect*, and thus the SA image positions highlighted in dark grey in Figure 3 do not contain usable information, being ignored in the selection process. To consider different solutions in terms of number of views, thus impacting complexity, and positioning, thus impacting the amount of disparity, the following SA image selection topologies have been defined:

**1) High-density SA image selection**: This topology considers a rather large number of SA images from the multi-view array, as illustrated in Figure 3 (a), where the selected SA images are highlighted in red. To arrange the selected SA images into a sequence, two different scanning orders are proposed: (i) *row-major scanning*, which concatenates SA images one row after another, from left to right, as illustrated in Figure 3 (b); and (ii) *snake-like scanning*, which also progresses row-wise, but the rows are alternatively scanned from left to right and right to left, as illustrated in Figure 3 (c).

**2) Max-disparity SA image selection**: This topology considers those SA images corresponding to the multi-view array's borders, thus considering the SA images for which the viewpoint changes the most and thus has the maximum disparity, as illustrated in Figure 3 (d). Some of the selected SA images may not be of the highest quality, due to the near vignetting effect.

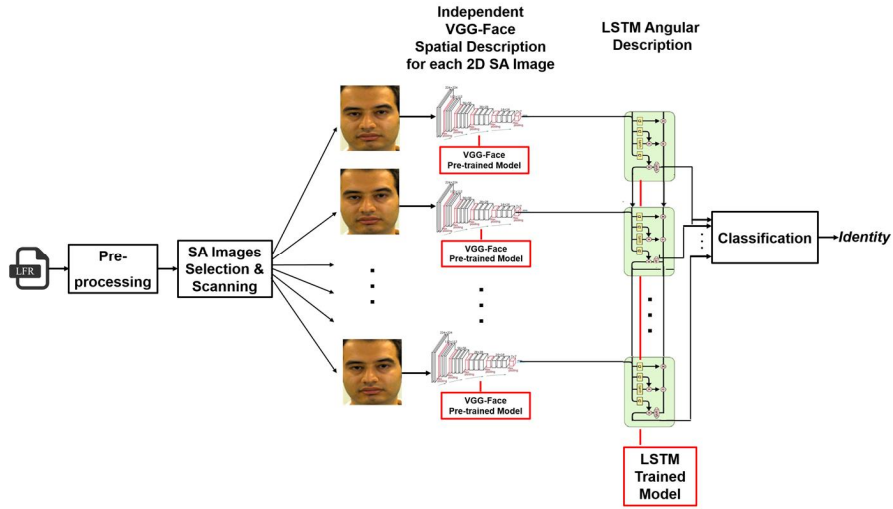

**Figure 2:** End-to-end architecture for the proposed double-deep learning framework.

**3) Mid-density SA image selection**: In this case, the selected SA images capture horizontal, vertical, and both horizontal and vertical parallaxes. The SA images considered are: (i) middle row – see Figure 3 (e); (ii) middle column – see Figure 3 (f); and (iii) combination of middle row and middle column – see Figure 3 (g). There are two possible ways to combine the horizontal and vertical angular information for the topology in Figure 3 (g): i) scanning the horizontal SA images followed by the vertical ones; or ii) processing each direction separately and then applying a sum rule score-level fusion, by adding the LSTM softmax classifier scores obtained for the horizontal and vertical SA images, as illustrated in Figure 4. It will be seen later that the performances for these two approaches may be rather different.

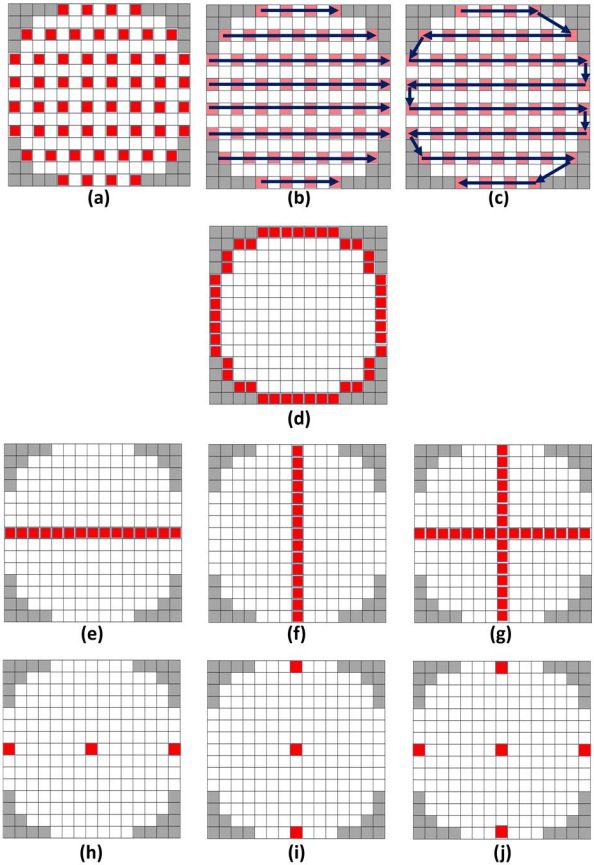

**Figure 3:** (a) High-density SA image selection topology; (b) row-major scanning order; (c) snake-like scanning order; (d) max-disparity SA image selection topology; (e) mid-density horizontal SA image selection topology; (f) mid-density vertical SA image selection topology; (g) mid-density horizontal and vertical SA image selection topology; (h) low-density horizontal SA image selection topology; (i) low-density vertical SA image selection topology; (j) low-density horizontal and vertical SA image selection topology.

**4) Low-density SA image selection:** Due to the considerable large computational power and memory resources required, exploiting the spatio-angular information from a considerable number of SA images may not always be the best option. Thus, a low-density sampling of the SA images should be also considered. Since results in [13] and [20] show a clear performance improvement for light field based face recognition and presentation attack detection as the SA images' disparity increases, the central view SA image along with two SA images at the maximum horizontal and vertical disparities from the central view are selected, as illustrated in Figure 3 (h) and Figure 3 (i), respectively. Figure 3 (j) shows the selection of both these horizontal and vertical SA images, for which the two combination approaches described above may be applied.

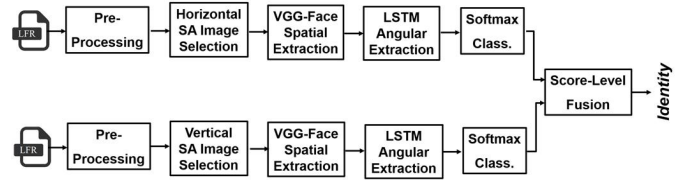

**Figure 4:** Score-level fusion for combining the horizontal and vertical inter-view, angular scores.

*D. VGG-Face Spatial Description*

The VGG-Face descriptions are obtained by running the VGG-16 CNN [23] without the last two fully connected layers, as demonstrated to be efficient in [22], thus including 13 convolutional layers, followed by one fully connected layer. The VGG-Face descriptor has been trained over 2.6 million face images, covering rich variations in expression, pose, occlusion, and illumination, obtaining a so-called *pre-trained VGG-Face model* for face recognition, containing 144 million weights.

In the proposed face recognition framework, the pre-trained VGG-Face model is used, thus implying no additional training at this stage. The VGG-Face descriptor is independently applied to each selected SA image. The output intra-view description is a fixed length feature vector, with a total of 4096 elements.

*E. LSTM Angular Description*

The VGG-Face descriptor only deals with the intra-view, spatial information within a 2D image. However, for a multi-view array of 2D SA images, it is possible to additionally exploit the inter-view, angular information available in the light field image to improve the face recognition performance.

Recurrent neural networks (RNN) can be used to extract higher dimensional dependencies from sequential data. The RNN units, called *cells*, have connections not only between the subsequent layers, but also into themselves, to keep information from previous inputs. To train a RNN, the so-called *back-propagation through time* algorithm can be used [62]. Simple RNN models can easily learn short-term dependencies. However, performing error back-propagation through long-term dependencies is a very challenging problem as gradients tend to vanish [63]. The Long Short-Term Memory (LSTM) RNN [24] is designed to learn both long- and short-term dependencies, using learned gating functions. LSTM has recently achieved impressive results on many large-scale learning tasks, such as speech recognition [64], language translation [65], activity recognition [58], and image captioning and video description [60]. Therefore, LSTM based networks are now widely used in many cutting-edge applications, notably Google Translate, Facebook, Siri or Amazon's Alexa.

An LSTM network is composed of cells whose outputs evolve through the network based on past memory content. Each LSTM cell has three inputs, including an input feature vector, an input hidden state coming from the previous cell, and a common cell state, which is shared between the cells. Each cell is controlled by three gates, *input, forget*, and *output* gates, allowing the network to decide when to forget the previous states and when to update the current state given new information. This structure allows LSTM to update the cell state and produces a hidden state as the output of the current cell and the input to the next cell. Applying this structure to a sequence of SA image descriptions (and not a sequence of images along time) enables the LSTM to learn a long short-term inter-view, angular model when using light field images for face recognition. The short-term and long-term dependencies can be learned from input hidden states coming from previous cells and the common LSTM cell state, respectively. The model obtained from the LSTM learning process can then be used for description creation during the testing phase.

The combination of VGG and LSTM networks has recently been used to learn spatio-temporal (video) models for different visual classification and description tasks. However, using the LSTM to learn the inter-view, angular model from a sequence of descriptions extracted from a set of SA images representing a light field image offers a novel approach, never tried before for any visual recognition task. As shown in Figure 2, the adopted LSTM network includes one LSTM cell with peephole connections per each selected SA image in the sequence. Based on the scanning order considered, the VGG-Face intra-view, spatial descriptions extracted from the SA images are passed to the corresponding LSTM cell. The output of each LSTM cell, corresponding to its hidden state, describes the short-term and long-term angular dependencies captured so far. The VGG+LSTM network has been trained with the MSE loss function, and batch normalization [66] has been used to control the distributions of feedforward network activations.

*F. LSTM Hyper-parameters*

LSTM has a number of hyper-parameters for model initialization whose optimization is of major importance for the final recognition performance, notably:

- **LSTM hidden layer size:** This hyper-parameter controls the size of the hidden layer in the LSTM units, which is also the size of each LSTM cell's output. A small hidden layer size requires setting fewer parameters, but it may lead to underfitting. A larger hidden layer size gives the network more capacity for convergence, while increasing the required training time. However, a too large hidden layer size may result in overfitting, thus highlighting the importance of appropriately adjusting the hidden layer size.
- **Batch size**: The input data can be divided into a number of batches, each used for one round of network weights update. There are two advantages of training a deep learning network using batches instead of the whole input data at once: i) decreasing the computational complexity, increasing the parallelization ability and needing less memory; and ii) performing a better training with stronger generalization ability as the network can escape from local minima [67] [68]. Nevertheless, it should be noted that a high number of batches, i.e., small batches, may lead to less accurate gradient estimation during the learning process.
- **Number of training epochs**: One epoch is a full forward-backward pass of all training samples through the network. Each epoch may consider a number of iterations, in case the whole data is divided into batches. The number of epochs should be selected in such a way that it guarantees network convergence within a reasonable training time.

*G. Softmax Classification*

The output (hidden state) of each LSTM cell is used as input to a softmax classifier and includes: i) short-term dependencies, corresponding to the recently observed view-point changes; and ii) long-term dependencies, corresponding to all the view-point changes observed so far (in the VGG-Face descriptions sequence). Then the average of the classification probabilities across the selected SA images, predicts the most probable label and thus the final output. This averaging mechanism, which has been widely used in the literature in the context of spatio-temporal frameworks for visual recognition tasks [60], considers all LSTM hidden states to exploit both the full short- and long-term angular dependencies. This approach offers a comprehensive angular description model for visual recognition. The alternative of only using the output of the last LSTM cell [59] may not exploit the full angular dependencies as the network tends to forget the long-term dependencies observed at the beginning, due to the LSTM *forget* gate mechanism; this would result into offering a slightly lower performance than the proposed solution.

IV. PERFORMANCE ASSESSMENT METHODOLOGY

This section presents the test material, experimental evaluation protocols, and the state-of-the-art recognition solutions considered for benchmarking purposes.

*A. Test material: IST-EURECOM LFFD*

The IST-EURECOM LFFD [28] is the first, and currently the only, available light field face database including the raw light field images. The IST-EURECOM LFFD consists of light field images captured by a Lytro ILLUM camera [36] from 100 subjects, including 20 images per person, per each of two acquisition sessions, with a temporal separation between 1 and 6 months. The face variations considered in IST-EURECOM LFFD (see illustration in Figure 5) can be categorized into 6 dimensions:

1. *Neutral*: one image captured with neutral emotion, standard illumination, and frontal pose;
2. *Emotions*: three images captured with happy, angry and surprise emotions;
3. *Actions*: two images captured with closed eyes and open mouth actions;
4. *Poses*: six images captured while looking up, looking down, right half-profile, right profile, left half-profile and left profile poses;
5. *Illumination*: two images with high and low illuminations;
6. *Occlusions*: six images with eye occluded by hand, mouth occluded by hand, glasses, sunglasses, mask and hat.

In this paper, all the light field images in the IST-EURECOM LFFD, are used to assess the face recognition performance; the corresponding SA images are cropped to the face area to minimize the impact of the background in the recognition process. This database corresponds to a practical scenario where the subjects present themselves to a fixed camera with a controlled background, but significant flexibility is allowed in terms of pose, expressions and occlusions. This is a rather common and realistic scenario in business and industrial environments where the facial images to be recognized are captured in, at least partly, constrained conditions.

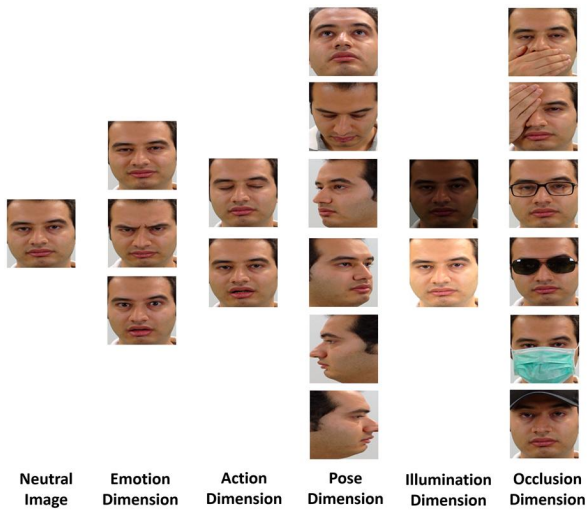

**Figure 5:** IST-EURECOM LFFD: Illustration of 2D central SA cropped images for one acquisition session.

*B. Evaluation Protocols*

To assess the overall performance of the proposed recognition solution as well as to study its sensitivity to the available training data, both in terms of number of training samples and facial variations, two evaluation protocols with practical meaningfulness are proposed. The definitions of the protocols are as follows:

- **Protocol 1**: The training set contains only the neutral light field images from the first acquisition session (1 light field image per subject), while the validation set includes the left and right half-profile images from the same acquisition session (2 images per subject), thus corresponding to a low-complexity training scenario; the testing set includes all the light field images from the second acquisition session (see Figure 6 (a)). This *'single training image per person'* protocol is the simplest protocol considered, but it is also the most challenging in terms of recognition performance.
- **Protocol 2**: The training set contains all twenty database face variations captured during the first acquisition session, while the validation and testing sets each consider half of the second session images (see Figure 6 (b)), thus corresponding to a higher complexity training scenario. This scenario is less challenging in terms of recognition performance as the recognition system learns more in the training phase.

The first protocol (*Protocol 1*) assumes a rather simple acquisition phase by considering only a single neutral-frontal image for training, and the left and right half-profile images from the same acquisition session for validation. This first protocol corresponds to a scenario where each person registers into the system by quickly taking 3 photos in a controlled setup, similar to the famous police station paradigm. Testing is done by considering all facial variations captured in the second acquisition session, assuming that the recognition should be robust to real-life conditions where the face may have expressions, be partly occluded, etc. In this case, the recognition system has not been exposed/trained to many of the facial variations with which it will be tested.

The second protocol (*Protocol 2*) assumes a more complex acquisition phase, considering more training images, under the assumption that the increased complexity will result in a better trained and thus more knowledgeable model, which should later offer a better recognition performance. This protocol divides the available database material into disjoint training (50%), validation (25%), and testing (25%) sets where the first session images are all used for training. In this case, the recognition system has been initially exposed/trained to more facial variations, investing on acquisition and training complexity to get later a better recognition performance.

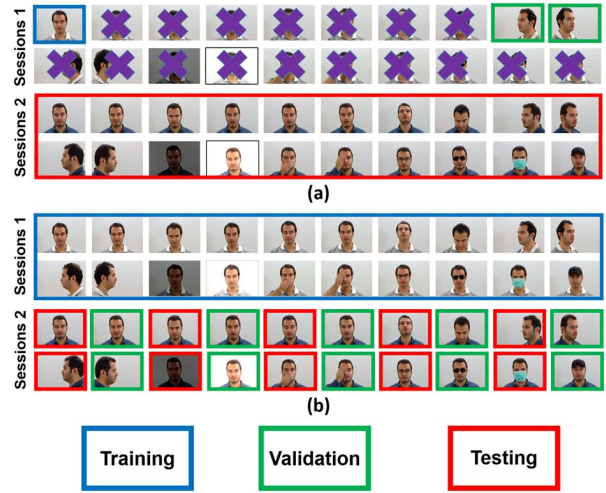

**Figure 6:** IST-EURECOM LFFD (non-cropped) database division into training, validation and testing sets for (a) Protocol 1; (b) Protocol 2.

The two protocols correspond to *cooperative user scenarios*, offering different trade-offs in terms of initial setting complexity and later recognition performance. The first protocol has multiple practical applications, such as in access control systems, where the users can be registered into the system by quickly taking a mugshot, including a frontal-view and two side-view photos in a controlled setup. Then, the goal is to recognize a person from an image captured at a different time in non-ideal conditions, e.g. exhibiting unpredictable facial variations. The second protocol corresponds to a more cooperative user scenario for usage in scenarios with increased security requirements, where the users are willing to cooperate during the registration phase, simulating different facial variations, over a range of expressions, actions, poses, illuminations, and occlusions, to capture as many variations as possible during the enrollment phase so that they can be more easily recognized during the daily operation of the system.

For both protocols, the training set is used to obtain the LSTM model weights, the validation set is used to tune the training model hyper-parameters and the testing set is used for the final system performance assessment. By considering a multi-label classification task (face recognition), at least one image from each subject (classes) with whom the system will be validated/tested must be available during the training stage. If a new subject is to be recognized, the database has to be extended with corresponding images and the classification model has to be re-trained (fine-tuned), as the new subject is an *unseen* label in the previous model. As the performance of the model being trained depends on a set of hyper-parameters, a disjoint set of validation samples are used to select the hyper-parameter values leading to the best performance. As usual in the literature, recognition rate at rank-1 is the metric adopted to report the results.

## C. Benchmarking Solutions

The competing recognition solutions considered for benchmarking purposes are grouped into three categories, summarized in Table 2:

1. **Conventional 2D solutions**, notably PCA [38], HOG [69], LBP [70], VGG [22], VGG+PCA [71] and VGG+ICA [72], which process **only** the (single) central view 2D SA image;
2. **Light field-limited solutions**, notably VGG-$D^3$ [15] and DLBP [73], which process **only** the light field central view data, notably using its texture and corresponding disparity and depth maps, whose computation may however require access to the complete light field data;
3. **Light field-full solutions**, notably MPCA [74], LFLBP [13], and HOG+HDG [14], which process **multiple or all** the light field views data, and thus need access to the full light field, targeting to exploit the spatio-angular light field information in the form of a multi-view SA array.

**Table 2:** Overview of benchmarking solutions.

| Ref. | Year | Type | Feature Extractor | Classifier(s) |
|---|---|---|---|---|
| [38] | 1991 | Conv. 2D | PCA | SVM |
| [69] | 2011 | Conv. 2D | HOG | SVM |
| [70] | 2012 | Conv. 2D | LBP | SVM |
| [22] | 2015 | Conv. 2D | VGG | SVM; k-NN (Manhattan and Euclidean distance metrics) |
| [71] | 2017 | Conv. 2D | VGG+PCA | SVM; k-NN (Manhattan and Euclidean distance metrics) |
| [72] | 2019 | Conv. 2D | VGG+ICA | SVM; k-NN (Manhattan and Euclidean distance metrics) |
| [73] | 2014 | LF-Limitted | DLBP | SVM |
| [15] | 2018 | LF-Limitted | VGG-$D^3$ | SVM |
| [74] | 2008 | LF-Full | MPCA | SVM |
| [13] | 2017 | LF-Full | LFLBP | SVM |
| [14] | 2018 | LF-Full | HOG+HDG | SVM |

In this paper, all the extracted face descriptions have been classified using a SVM classifier. For the VGG, VGG+PCA, and VGG+ICA the descriptions are also classified using k-NN with Manhattan ($L^1$) and Euclidean ($L^2$) distance metrics. The combination of these three VGG-based feature extraction solutions with the three classifiers leads to nine VG-based baseline solutions, which are appropriate for comparison with the proposed VGG+LSTM recognition framework. To obtain results for the solutions classified as 'Light field-limited solutions', the disparity maps were obtained using the methods proposed in [75] and [76], and the depth maps using the method proposed in [77]. To apply the new test protocols proposed in this paper, all the 17 benchmarking solutions considered were re-implemented by the authors of this paper and performance results were obtained considering the best parameter settings reported in the relevant original papers.

## V. PERFORMANCE ASSESSMENT

This section reports the performance assessment results obtained for the two proposed protocols. It starts by evaluating the impact of the hyper-parameter settings, notably analyzing the influence of the LSTM hidden layer size, the batch size, and the number of epochs to consider for network convergence. After, the impact of the various proposed SA image selection topologies and scanning methods is evaluated in terms of recognition accuracy and learning time. Once the optimal recognition framework configuration is decided, comprehensive comparisons, in terms of recognition accuracy, processing time, and memory efficiency are performed between the proposed recognition framework and relevant alternative recognition solutions.

### A. Hyper-Parameter Evaluation: Hidden Layer Size

The study of recognition performance sensitivity to the size of the LSTM hidden layers is reported first. Figure 7 illustrates the rank-1 recognition performance for hidden layer sizes of 32, 64, 128, 256, and 512 for Protocol 1 (Figure 7 (a)) and Protocol 2 (Figure 7 (b)) validation sets, after training with all the proposed SA image selection methods. These results are reported considering a batch size of 34 and 667 (1/3 of the input data), respectively for protocols 1 and 2, and 50 epochs. These values were selected after some initial experimentation, which showed the suitability of these values for network initialization.

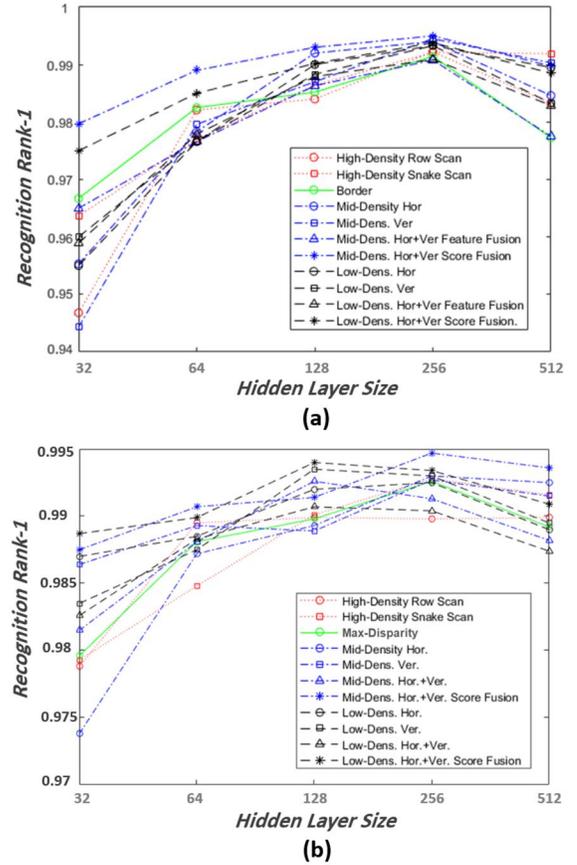

**Figure 7:** Recognition rank-1 versus hidden layer size considering all proposed SA image selection methods for (a) Protocol 1 and (b) Protocol 2.

The results show a clear improvement on the recognition performance as the hidden layer size is increased up to 256. The recognition accuracy is not further increased by considering larger LSTM hidden layer sizes, even gradually decreasing for a size of 512. This may be due to overfitting and shows that LSTM tends to converge to a complex model that is not well captured using a too small hidden layer size.

## B. Hyper-Parameter Evaluation: Batch Size

In theory, the batch size should be adjusted to have an accurate gradient estimation while avoiding overfitting. Figure 8 illustrates the recognition performance for protocols 1 and 2 validation sets, when considering between 2 and 6 batches, resulting in batch sizes of 50, 34, 25, 20 and 17 for Protocol 1, and 1000, 667, 500, 400 and 333 for Protocol 2. Results are reported for 50 epochs, after setting the hidden layer size to 256, the best size identified in Section IV-A.

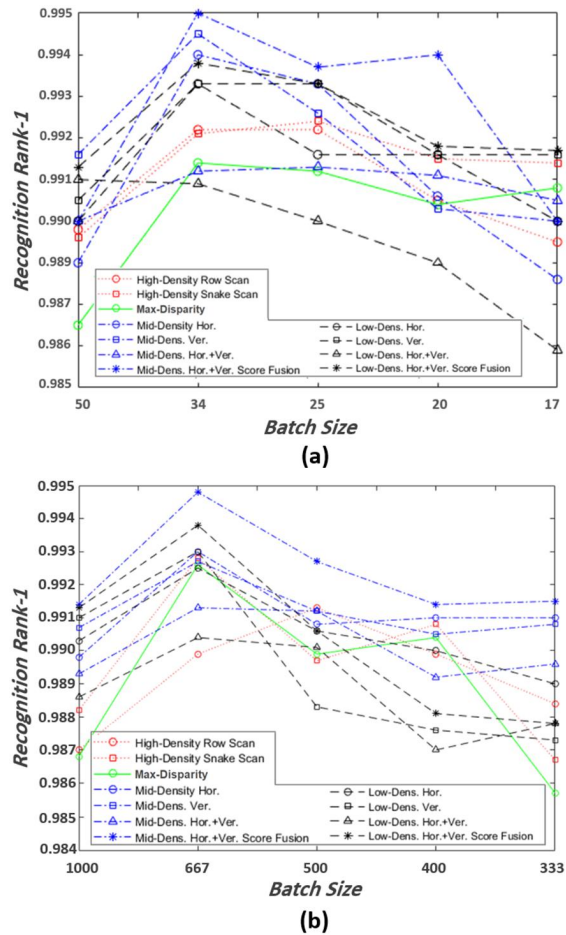

**Figure 8:** Recognition rank-1 versus number of batches considering all proposed SA image selection methods for (a) Protocol 1 and (b) Protocol 2.

The results presented in Figure 8 show that using three batches, i.e., batch sizes of 34 and 667, respectively for protocols 1 and 2, allows a good gradient estimation, leading to the best recognition performance for almost all cases. It should be noted that since the LSTM inputs are VGG face descriptions, the input dimension is very small, i.e., 4096, thus justifying the better performance obtained by the large batch size selected for Protocol 2. It is also possible to observe that mid-density SA image selection methods are more robust to changes in the number of batches, when compared to the other SA image selection methods.

## C. Hyper-Parameter Evaluation: Number of Training Epochs

The number of training epochs, which directly impacts the required training time, should be minimized while guaranteeing network convergence. Figure 9 shows the recognition performance for the Protocol 1 (Figure 9 (a)) and Protocol 2 (Figure 9 (b)) validation sets when varying the number of training epochs, after training with all the proposed SA image selection methods. Results are reported by setting the hidden layer size and the number of batches to 256 and 3, respectively, based on the conclusions from the previous sub-sections.

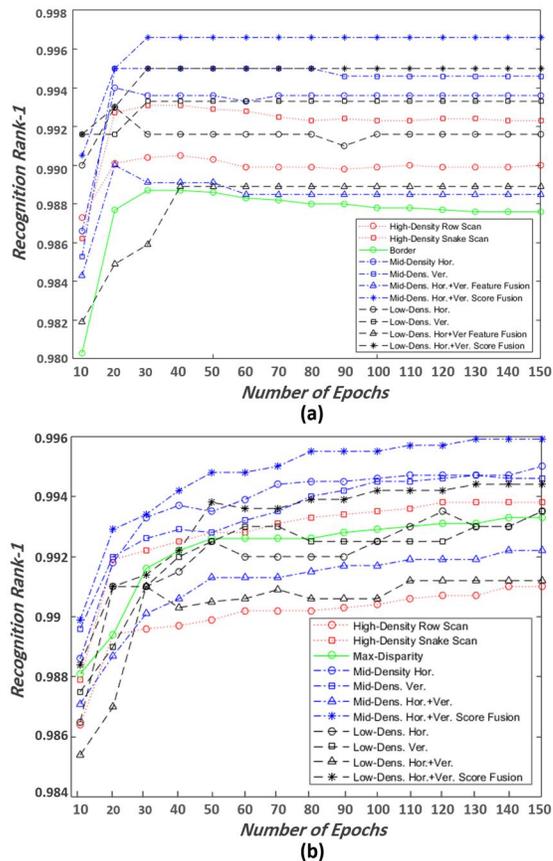

**Figure 9:** Recognition rank-1 versus number of training epochs considering all proposed SA image selection methods for (a) Protocol 1 and (b) Protocol 2.

The experimental results show that considering 40 and 130 training epochs, respectively for protocols 1 and 2, leads to a stable recognition performance for almost all the cases. The recognition performance remains almost constant for a higher number of epochs. The network converges much faster in Protocol 1 as the validation data is smaller. Hence, to keep a good trade-off between accuracy and training time and also to keep the same framework configuration for both evaluation protocols, the number of training epochs selected is 130.

## D. SA Image Selection Evaluation

As shown in Figure 9, for the high density SA image selection, the snake-like scanning offers superior performance over the row-major scanning, as it avoids the significant viewpoint feature discontinuities resulting from moving from the right-most SA image in a row to the left-most SA image in the next row.

It is also clear from Figure 9 that the mid-density SA image selection solution, capturing full angular information along the horizontal and vertical directions, achieves better average performance when compared to the high- and low-density selection methods. Among the proposed mid-density selection alternatives, the score-level fusion of horizontal and vertical angular models leads to the best performance. The alternative of performing a single combined scan implies a viewpoint feature discontinuity when moving from the last horizontal SA image (middle row) to the first vertical SA image (top row) which leads to a worse performance.

Table 3 shows the LSTM learning times (in seconds) for the different SA image selection methods for Protocol 2, with a total of 2000 training light field images, based on the best values adopted for the hyper-parameters. Additionally, Table 3 shows the running time for the validation step, considering 1000 light field images. Due to space constraints, complexity results are presented only for Protocol 2 which is the most critical as it considers the larger number of images during the training and validation phases.

**Table 3:** Learning times for the proposed SA image selection methods for Protocol 2 using 2000 training light field images.

| SA image selection method | VGG-Face extrac. time (s) | LSTM lear. time(s) | Total learn. time(s) | Learn. time per LF image (s) |
|---|---|---|---|---|
| High Dens. Row Scan | 1562 | 1815 | 3377 | 1.69 |
| High Dens. Snake Scan | 1562 | 1793 | 3355 | 1.68 |
| Corner | 1533 | 1811 | 3344 | 1.67 |
| Mid Dens. Hor. | 433 | 548 | 981 | 0.49 |
| Mid Dens. Ver. | 433 | 548 | 981 | 0.49 |
| Mid-Dens. Hor.+Ver. | 867 | 1085 | 1952 | 0.98 |
| **Mid-Dens. Hor.+Ver. Score** | **867** | **1106** | **1973** | **0.99** |
| Low Dens. Hor. | 173 | 111 | 284 | 0.14 |
| Low Dens. Ver. | 173 | 111 | 284 | 0.14 |
| Low-Dens. Hor.+Ver | 347 | 138 | 485 | 0.24 |
| Low-Dens. Hor.+Ver. Score | 347 | 231 | 578 | 0.29 |

**Table 4:** Validation times for the proposed SA image selection methods for Protocol 2 using 1000 validation light field images.

| SA image selection method | VGG-Face ext. time (s) | LSTM ext. time (s) | Total valid. time(s) | Valid. time per LF image (s) |
|---|---|---|---|---|
| High Dens. Row Scan | 781 | 3 | 784 | 0.78 |
| High Dens. Snake Scan | 781 | 3 | 784 | 0.78 |
| Corner | 767 | 3 | 770 | 0.77 |
| Mid Dens. Hor. | 217 | 1 | 218 | 0.22 |
| Mid Dens. Ver. | 217 | 1 | 218 | 0.22 |
| Mid-Dens. Hor.+Ver. | 434 | 2 | 436 | 0.44 |
| **Mid-Dens. Hor.+Ver. Score** | **434** | **2** | **436** | **0.44** |
| Low Dens. Hor. | 87 | 1 | 88 | 0.09 |
| Low Dens. Ver. | 87 | 1 | 88 | 0.09 |
| Low-Dens. Hor.+Ver | 174 | 1 | 175 | 0.18 |
| Low-Dens. Hor.+Ver. Score | 781 | 1 | 782 | 0.78 |

Table 3 and Table 4 include learning and validation times for each light field image, thus providing an estimate of the running time for other protocols, notably Protocol 1. Time measurements were performed on a standard 64-bit Intel PC with a Core i7 2.60 GHz processor, a 3 GB GeForce GTX 1060 graphics card, and 16 GB RAM, running Ubuntu 16.04.4 LTS. Considering the best performing SA image selection method, i.e., the mid-density horizontal and vertical with score-level fusion, the average learning and validation times per light field image are less than one second and half a second, respectively. This shows that the proposed recognition framework is a rather fast solution that can operate in (near) real-time, with most of the processing time being consumed by the VGG-Face descriptor.

Based on the validation experiments described so far, in terms of accuracy, network complexity, convergence speed, and the required learning and testing times, the best configuration for the proposed double-deep learning recognition framework, for both evaluation protocols, to be used from this point on for final recognition performance assessment, is summarized in Table 5.

**Table 5:** Selected configuration for the proposed light field based double-deep learning face recognition framework.

| LSTM hidden layer size | 256 |
|---|---|
| Number of batches | 1/3 of the input data |
| Number of training epochs | 130 |
| SA image selection method | Mid-density horizontal and vertical, with score-level fusion |

*E. Comparative Recognition Performance*

Tables 6 and 7 report the rank-1 recognition rates obtained, respectively, for test protocols 1 and 2, for the proposed double-deep learning framework and the seventeen benchmarking recognition solutions introduced in Section IV-C. It is important to keep in mind that the two protocols use different LSTM models, as they are derived using different training sets. The results in these tables are presented for the five recognition tasks corresponding to the LFFD database dimensions, and the best results are highlighted in bold.

**Comparison with conventional 2D recognition solutions:** The results clearly show that the proposed recognition framework performs considerably better than all tested 2D conventional face recognition solutions, including PCA [38], HOG [69], and LBP [70]. This is due to: i) adoption of a double-deep learning framework; and ii) exploitation of the spatio-angular information available in light field images. The proposed framework achieves average performance gains of 5.47%, 5.27%, and 5.72%, respectively, when compared to the baseline VGG-face, VGG-face+PCA, and VGG-face+ICA solutions using the best performing classifiers [22].

**Comparison with light field-limited solutions:** The obtained rank-1 recognition results also show that the proposed face recognition solution achieves better performance than all the considered light field-limited solutions [15] [73], for all face recognition tasks/protocols considered. Performance comparison of the proposed framework regarding the best performing fused deep VGG representation shows a 5.22% improvement, on average, for rank-1 recognition results.

**Comparison with light field-full solutions:** The obtained rank-1 recognition results demonstrate the superiority of the proposed framework when compared to the MPCA [74], LFLBP [13] and HOG+HDG [14] light field-full solutions. This is due to the double-deep learning of intra-view, spatial and inter-view, angular models from the light field images performed by the proposed framework.

The average rank-1 recognition rates obtained show that the proposed recognition framework is less sensitive to the number of training samples and the presence of facial variations in the training set, when compared to the benchmarking solutions.

The much-improved face recognition results under illumination variations illustrate the robustness of the proposed framework to illumination changes, highlighting the importance of exploiting the inter-view, angular variations, which are invariant to the intensity changes resulting from different illumination levels incorporated in the data acquisition process. The same is true for occlusions. Naturally, the less impressive result, although still with larger gains regarding the benchmarks, happens for the pose variations as the testing is made for right and left profile images after a training with only one frontal face, which justifies the low rank-1 recognition rates obtained for the pose variation in Protocol 1.

*F. Added Value of Light Field Information for Face Recognition*

The results presented in Tables 8 and 9 show the average rank-1 recognition results for all recognition tasks for some 2D baseline solutions against their corresponding light field based variants, e.g. PCA [48] against MPCA [52], respectively for test protocols 1 and 2. The average recognition gain clearly shows the added value of light field information for face recognition purposes. Considering that the 2D baseline VGG solution already achieves a very good performance, the significant average performance gains obtained again highlight the effectiveness of the proposed framework, which can additionally explore the inter-view, angular information by using an LSTM network that takes as input the VGG descriptions computed for each selected viewpoint.

*G. Processing Time and Compactness Analysis*

This section assesses the computational complexity and memory efficiency (compactness) for the proposed recognition framework and selected benchmarking solutions. The computational complexity is assessed by the processing/execution time measured on a standard 64-bit Intel PC with a 3.40 GHz processor, 16 GB RAM, and a GeForce GTX 1060 graphics card; MATLAB R2015b on Windows 10 was used for non-deep learning based solutions and PyTorch with CUDA 8 toolkit on UBUNTU 16.04 was used for deep learning based solutions.

To assess the computational complexity, Table 10 shows: i) the training times (in seconds) for the spatial description, (spatio-)angular description, classification, and total training; and ii) the testing times (in seconds) for the spatial description, (spatio-)angular description, classification, and total training, for the proposed solution and the seventeen benchmarking recognition solutions per each 2D/light field image. Naturally, there are no angular description times for non-light field based (2D) solutions. This table also summarizes the final description size in terms of number of elements and memory size in bytes for the various solutions as these performance metrics are relevant to analyze memory efficiency. The time and compactness metrics have been measured for each input image, either 2D or light field.

As can be observed from Table 10, although the total training and testing times for the proposed solution are higher than for some benchmarking solutions, this is expectable and the 'price' to pay for the additional exploitation of the angular information; still the total testing time per image is around half a second, thus facilitating operation in 'real-time'. Moreover, since the VGG spatial description process from multiple SA images has the largest impact on the overall framework complexity, the training and testing times can be significantly reduced by extracting the VGG spatial descriptions in parallel, which is possible with most modern processors and compilers.

The proposed framework also offers relatively compact representations when compared with the benchmarking solutions, thus simplifying the storage, retrieval, and transmission of the created spatio-angular descriptions. The compactness of the VGG+LSTM descriptions as the input to the low complexity Softmax classifier lead to the lowest classification time for the proposed solution during the testing phase in comparison with the benchmarking solutions. Finally, it is worth mentioning that the very high angular description times for [73] and [15] are related to the extraction of the disparity and depth maps, which are computationally very expensive.

**Table 6:** Protocol 1 assessment: Face recognition rank-1 for the proposed recognition framework and benchmarking solutions (maximum values in bold).

| Recognition Solution | | | | | Recognition task | | | | | |
|---|---|---|---|---|---|---|---|---|---|---|
| *Ref.* | *Year* | *Type* | *Descriptor* | *Classifier* | Neutral &Emotion | Action | Pose | Illumination | Occlusion | Average |
| [38] | 1991 | Conv. 2D | PCA | SVM | 28.50% | 28.00% | 06.67% | 12.50% | 16.33% | 17.40% |
| [69] | 2011 | Conv. 2D | HOG | SVM | 57.50% | 58.50% | 09.83% | 48.00% | 38.33% | 36.60% |
| [70] | 2012 | Conv. 2D | LBP | SVM | 16.75% | 18.50% | 06.67% | 12.00% | 09.33% | 11.20% |
| [22] | 2015 | Conv. 2D | VGG-Face | SVM | 99.50% | 99.00% | 56.33% | 99.00% | 74.67% | 79.00% |
| [22] | 2015 | Conv. 2D | VGG-Face | k-NN (Manhattan) | 98.50% | 98.00% | 54.16% | 98.00% | 73.00% | 77.45% |
| [22] | 2015 | Conv. 2D | VGG-Face | k-NN (Euclidean) | 98.50% | 98.00% | 55.66% | 98.00% | 73.83% | 78.15% |
| [71] | 2017 | Conv. 2D | VGG-Face + PCA | SVM | 99.25% | **99.50%** | 56.66% | **99.50%** | 75.50% | 79.40% |
| [71] | 2017 | Conv. 2D | VGG-Face + PCA | k-NN (Manhattan) | 98.75% | 98.50% | 54.66% | 98.00% | 70.50% | 76.95% |
| [71] | 2017 | Conv. 2D | VGG-Face + PCA | k-NN (Euclidean) | 98.25% | 98.50% | 56.00% | 98.50% | 74.66% | 78.55% |
| [72] | 2019 | Conv. 2D | VGG-Face + ICA | SVM | 99.25% | 98.50% | 57.66% | 99.00% | 78.00% | 80.30% |
| [72] | 2019 | Conv. 2D | VGG-Face + ICA | k-NN (Manhattan) | 97.75% | 97.00% | 54.33% | 97.50% | 75.00% | 77.80% |
| [72] | 2019 | Conv. 2D | VGG-Face + ICA | k-NN (Euclidean) | 98.25% | 97.50% | 57.16% | 98.00% | 77.16% | 79.50% |
| [73] | 2014 | LF-Limitted | DLBP | SVM | 59.25% | 64.50% | 30.33% | 24.50% | 22.33% | 36.55% |
| [15] | 2018 | LF-Limitted | VGG-D$^3$ | SVM | 99.50% | 99.00% | 56.50% | 99.00% | 75.50% | 79.50% |
| [74] | 2008 | LF-Full | MPCA | SVM | 36.75% | 33.50% | 07.50% | 14.50% | 19.67% | 20.30% |
| [13] | 2017 | LF-Full | LFLBP | SVM | 34.25% | 31.00% | 10.17% | 17.00% | 13.17% | 18.65% |
| [14] | 2018 | LF-Full | HOG+HDG | SVM | 62.25% | 62.50% | 12.00% | 62.00% | 41.33% | 40.90% |
| Prop. Solution | | LF-Full | Proposed VGG+LSTM | Softmax | **100%** | **99.50%** | **71.00%** | 99.00% | **92.17%** | **88.75%** |

**Table 7:** Protocol 2 assessment: Face recognition rank-1 for the proposed recognition framework and benchmarking solutions (maximum values in bold).

| Recognition solution | | | | | Recognition task | | | | | |
|---|---|---|---|---|---|---|---|---|---|---|
| Ref. | Year | Type | Descriptor | Classifier | Neutral &Emotion | Action | Pose | Illumination | Occlusion | Average |
| [38] | 1991 | Conv. 2D | PCA | SVM | 48.50% | 45.00% | 30.67% | 23.00% | 39.00% | 37.40% |
| [69] | 2011 | Conv. 2D | HOG | SVM | 81.00% | 75.00% | 22.33% | 86.00% | 61.67% | 57.50% |
| [70] | 2012 | Conv. 2D | LBP | SVM | 18.50% | 18.00% | 18.67% | 27.00% | 18.67% | 19.40% |
| [22] | 2015 | Conv. 2D | VGG-Face | SVM | 98.50% | 99.00% | 95.67% | 99.00% | 97.00% | 97.30% |
| [22] | 2015 | Conv. 2D | VGG-Face | k-NN (Manhattan) | 99.50% | **100%** | 93.33% | **100%** | **98.67%** | 97.50% |
| [22] | 2015 | Conv. 2D | VGG-Face | k-NN (Euclidean) | 99.50% | **100%** | 93.00% | **100%** | **98.67%** | 97.40% |
| [71] | 2017 | Conv. 2D | VGG-Face + PCA | SVM | 99.00% | 99.00% | 93.00% | 99.0% | 98.00% | 96.90% |
| [71] | 2017 | Conv. 2D | VGG-Face + PCA | k-NN (Manhattan) | 99.5% | **100%** | 91.67% | **100%** | 98.33% | 96.90% |
| [71] | 2017 | Conv. 2D | VGG-Face + PCA | k-NN (Euclidean) | 99.50% | **100%** | 93.00% | **100%** | 98.33% | 97.30% |
| [72] | 2019 | Conv. 2D | VGG-Face + ICA | SVM | 97.50% | 99.00% | 91.33% | 96.00% | 96.67% | 95.40% |
| [72] | 2019 | Conv. 2D | VGG-Face + ICA | k-NN (Manhattan) | 98.00% | 99.00% | 92.33% | 96.00% | 95.33% | 95.40% |
| [72] | 2019 | Conv. 2D | VGG-Face + ICA | k-NN (Euclidean) | 97.50% | 99.00% | 92.66% | 97.00% | 95.33% | 95.50% |
| [73] | 2014 | LF-Limitted | DLBP | SVM | 88.50% | 89.00% | 67.00% | 51.00% | 56.67% | 68.80% |
| [15] | 2018 | LF-Limitted | VGG-D³ | SVM | 99.50% | 99.00% | 95.00% | 100% | 96.67% | 97.30% |
| [74] | 2008 | LF-Full | MPCA | SVM | 48.00% | 51.00% | 29.67% | 41.00% | 38.00% | 39.10% |
| [13] | 2017 | LF-Full | LFLBP | SVM | 39.00% | 34.00% | 26.00% | 36.00% | 30.67% | 31.80% |
| [14] | 2018 | LF-Full | HOG+HDG | SVM | 83.00% | 81.00% | 33.67% | 84.00% | 65.67% | 62.90% |
| Prop. Solution | | LF-Full | Proposed VGG+LSTM | Softmax | **100%** | **100%** | **96.33%** | **100%** | **98.67%** | **98.50%** |

**Table 8:** Protocol 1 average rank-1 recognition results for some conventional 2D baseline solutions against their light field based variants (maximum values in bold).

| Recognition solution | | Average performance for all recognition tasks | | |
|---|---|---|---|---|
| Conv. 2D | LF based | 2D average | LF based average | Gain |
| VGG-Face [22] | Proposed VGG+LSTM | 79.00% | 88.75% | **9.25%** |
| PCA [38] | MPCA [74] | 20.30% | 17.40% | 2.90% |
| LBP [70] | LFLBP [13] | 18.65% | 11.20% | 7.45% |
| HOG [69] | HOG+HDG [14] | 40.90% | 36.60% | 4.30% |

**Table 9:** Protocol 2 average rank-1 recognition results for some conventional 2D baseline solutions against their light field based variants (maximum gain in bold).

| Recognition solution | | Average performance for all recognition tasks | | |
|---|---|---|---|---|
| Conv. 2D | LF based | 2D average | LF based average | Gain |
| VGG-Face [22] | Proposed VGG+LSTM | 97.30% | 98.50% | 1.20% |
| PCA [38] | MPCA [74] | 37.40% | 39.10% | 1.70% |
| LBP [70] | LFLBP [13] | 19.40% | 31.80% | **12.40%** |
| HOG [69] | HOG+HDG [14] | 57.50% | 62.50% | 5.00% |

**Table 10:** Processing time and compactness performance for the proposed recognition framework and benchmarking solutions per each 2D/light field image (Time values are in in seconds and minimum values in bold).

| Recognition solution | | | Training time / image | | | | Testing time / image | | | | Description size | |
|---|---|---|---|---|---|---|---|---|---|---|---|---|
| Descriptor | Classifier | Type | Spatial description | (Spatio-)angular description | Classification | Total | Spatial description | (Spatio-)angular description | Classification | Total | Number of elements | Bytes |
| PCA [38] | SVM | 2D | 0.149 | N/A | 0.025 | 0.174 | 0.149 | N/A | 0.018 | 0.149 | 4,095 | 1764 |
| HOG [69] | SVM | 2D | 0.107 | N/A | 0.070 | 0.177 | 0.107 | N/A | 0.040 | 0.107 | 8,100 | 60773 |
| LBP [70] | SVM | 2D | **0.013** | N/A | 0.040 | 0.053 | **0.013** | N/A | 0.030 | **0.013** | 4,096 | 1801 |
| VGG-Face [22] | SVM | 2D | 0.016 | N/A | 0.030 | 0.046 | 0.016 | N/A | 0.020 | 0.016 | 4,096 | 15473 |
| VGG-Face [22] | k-NN (Manh.) | 2D | 0.016 | N/A | 0.024 | 0.040 | 0.016 | N/A | 0.017 | 0.016 | 4,096 | 15473 |
| VGG-Face [22] | k-NN (Eucl.) | 2D | 0.016 | N/A | 0.024 | 0.040 | 0.016 | N/A | 0.017 | 0.016 | 4,096 | 15473 |
| VGG-Face + PCA [71] | SVM | 2D | 0.022 | N/A | 0.009 | 0.031 | 0.019 | N/A | 0.005 | 0.019 | 512 | 492 |
| VGG-Face + PCA [71] | k-NN (Manh.) | 2D | 0.022 | N/A | 0.008 | 0.030 | 0.019 | N/A | 0.004 | 0.019 | 512 | 492 |
| VGG-Face + PCA [71] | k-NN (Eucl.) | 2D | 0.022 | N/A | 0.008 | 0.030 | 0.019 | N/A | 0.004 | 0.019 | 512 | 492 |
| VGG-Face + ICA [72] | SVM | 2D | 0.020 | N/A | 0.007 | 0.027 | 0.019 | N/A | 0.004 | 0.019 | **256** | **179** |
| VGG-Face + ICA [72] | k-NN (Manh.) | 2D | 0.020 | N/A | **0.006** | 0.026 | 0.019 | N/A | 0.003 | 0.019 | **256** | **179** |
| VGG-Face + ICA [72] | k-NN (Eucl.) | 2D | 0.020 | N/A | **0.006** | 0.026 | 0.019 | N/A | 0.003 | 0.019 | **256** | **179** |
| DLBP [73] | SVM | LF | 0.026 | 391.061 | 0.492 | 391.57 | 0.026 | 391.061 | 0.271 | 391.08 | 73,728 | 11116 |
| VGG-D³ [15] | SVM | LF | 0.048 | 397.473 | 0.150 | 397.67 | 0.048 | 397.473 | 0.100 | 397.52 | 16,384 | 28911 |
| MPCA [73] | SVM | LF | 0.149 | 0.141 | 0.029 | 0.319 | 0.149 | 0.141 | 0.022 | 0.290 | 4,095 | 1764 |
| LFLBP [13] | SVM | LF | **0.013** | 0.023 | 0.430 | 0.466 | **0.013** | 0.023 | 0.240 | 0.036 | 65,536 | 7581 |
| HOG+HDG [14] | SVM | LF | 0.302 | 0.195 | 0.170 | 0.667 | 0.302 | 0.195 | 0.080 | 0.497 | 16,200 | 122134 |
| Proposed VGG+LSTM | Softmax | LF | 0.480 | 0.295 | 0.033 | 0.808 | 0.480 | **0.031** | **0.002** | 0.511 | 7680 | 29011 |

## VI. Summary and Future Work

This paper proposes a double-deep learning framework for light field based face recognition. The proposed learning framework includes for the first time an LSTM network to learn a model expressing the inter-view, angular information present in the multiple viewpoint SA images extracted a light field image. A comprehensive set of experiments, in terms of accuracy, network complexity, convergence speed, and learning and testing times for two evaluation protocols, has been conducted with the IST-EURECOM light field face database. The results show that the proposed solution achieves superior recognition performance over nine state-of-the-art benchmarking solutions.

The proposed spatio-angular framework deals with inter-view angular information as a descriptions sequence. However, as the position of each SA image within the multi-view array and the SA images scanning order are known, there is some additional information about the inter-view angular information/ dependencies, such as parallax, that could be further exploited during the learning to increase the recognition accuracy and/or convergence speed. An extension of the LSTM for spatio-angular visual recognition tasks, further exploiting the additional angular information, will be considered as future work.

## VII. Acknowledgment

This work has been partially supported by Instituto de Telecomunicações under Fundação para a Ciência e Tecnologia, Grant UID/EEA/ 50008/2013. Authors also acknowledge support from European Cooperation in Science and Technology (COST) framework under COST Action CA16101- MULTI-FORESEE.

## References

[1] A. Jain, K. Nandakumarb and A. Ross, "50 years of biometric research: Accomplishments, challenges, and opportunities," *Pattern Recognition Letters,* vol. 79, no. 1, pp. 80-105, Aug. 2016.

[2] M. Günther, L. El Shafey and S. Marcel, "2D face recognition: An experimental and reproducible research survey," Technical Report Idiap-RR-13, Martigny, Switzerland, Apr. 2017.

[3] A. Sepas-Moghaddam, F. Pereira and P. Correia, "Face recognition: A novel multi-level taxonomy based survey," arXiv:1901.00713, Jan. 2019.

[4] G. Hu, Y. Yang, D. Yi, J. Kittler, W. Christmas, S. Li and T. Hospedales, "When face recognition meets with deep learning: An Evaluation of convolutional neural networks for face recognition," in *International Conference on Computer Vision Workshop*, Santiago, Chile, Dec. 2015.

[5] M. Wang and W. Deng, "Deep face recognition: A survey," arXiv:1804.06655, Feb. 2019.

[6] L. Best-Rowden, H. Han, C. Otto, B. Klare and A. Jain, "Unconstrained face recognition: Identifying a person of interest from a media collection," *IEEE Transactions on Information Forensics and Security,* vol. 9, no. 12, pp. 2144-2157, Dec. 2014.

[7] R. Ng, M. Levoy, M. Bradif, G. Duval, M. Horowitz and P. Hanrahan, "Light field photography with a hand-held plenoptic camera," Tech Report CSTR 2005-02, Stanford, CA, USA, Feb. 2005.

[8] M. Levoy and P. Hanrahan, "Light field rendering," in *23rd Annual Conference on Computer Graphics and Interactive Techniques*, New York, NY, USA, Aug. 1996.

[9] R. Raghavendra, B. Yang, K. Raja and C. Busch, "A new perspective-Face recognition with light-field camera," in *International Conference on Biometrics*, Madrid, Spain, Jun. 2013.

[10] R. Raghavendra, K. Raja, B. Yang and C. Busch, "Multi-face recognition at a distance using light-field camera," in *International Conference on Intelligent Information Hiding and Multimedia Signal Processing*, Beijing, China, Jul. 2013.

[11] R. Raghavendra, K. Raja, B. Yang and C. Busch, "Comparative evaluation of super-resolution techniques for multi-face recognition using light-field camera," in *International Conference on Digital Signal Processing*, Santorini, Greece, Jul. 2013.

[12] R. Raghavendra,, K. Raja and C. Busch, "Exploring the usefulness of light field cameras for biometrics: An empirical study on face and iris recognition," *IEEE Transaction on Information Forensics and Security,* vol. 11, no. 5, pp. 922-936, May 2016.

[13] A. Sepas-Moghaddam, P. Correia and F. Pereira, "Light field local binary patterns description for face recognition," in *International Conference on Image Processing*, Beijing, China, Sep. 2017.

[14] A. Sepas-Moghaddam, F. Pereira and P. Correia, "Ear recognition in a light field imaging framework: A new perspective," *IET Biometrics,* vol. 7, no. 3, p. 224–231, May 2018.

[15] A. Sepas-Moghaddam, P. Correia, K. Nasrollahi, T. Moeslund and F. Pereira, "Light field based face recognition via a fused deep representation," in *International Workshop on Machine Learning for Signal Processing*, Aalborg, Denmark, Sep. 2018.

[16] S. Kim, Y. Ben and S. Lee, "Face liveness detection using a light field camera," *Sensors,* vol. 14, no. 12, pp. 71-99, Nov. 2014.

[17] R. Raghavendra, K. Raja and C. Busch, "Presentation attack detection for face recognition using light field camera," *IEEE Transactions on Image Processing,* vol. 24, no. 3, pp. 1060-1075, Mar. 2015.

[18] Z. Ji, H. Zhu and Q. Wang, "LFHOG: A discriminative descriptor for live face detection from light field image," in *International Conference on Image Processing*, Phoenix, AZ, USA, Sep. 2016.

[19] A. Sepas-Moghaddam, L. Malhadas, P. Correia and F. Pereira, "Face spoofing detection using a light field imaging framework," *IET Biometrics,* vol. 7, no. 1, pp. 39-48, Jan. 2018.

[20] A. Sepas-Moghaddam, F. Pereira and P. Correia, "Light field based face presentation attack detection: Reviewing, benchmarking and one step further," *IEEE Transactions on Information Forensics and Security,* vol. 13, no. 7, pp. 1696-1709, Jul. 2018.

[21] A. Sepas-Moghaddam, F. Pereira and P. Correia, "Ear presentation attack detection: Benchmarking study with first lenslet light field database," in *European Signal Processing Conference*, Rome, Italy, Sep. 2018.

[22] O. Parkhi, A. Vedaldi and A. Zisserman, "Deep face recognition," in *British Machine Vision Conference*, Swansea, UK, Sep. 2015.

[23] K. Simonyan and A. Zisserman, "Very deep convolutional networks for large-scale image recognition," arXiv: 1409.1556, Apr. 2015.

[24] S. Hochreiter and J. Schmidhuber, "Long short-term memory," *Neural Computation,* vol. 9, no. 8, pp. 1735-1780, Nov. 1997.

[25] X. Shu, j. Tang, H. Lai, L. Liu and S. Yan, "Personalized age progression with aging dictionary," in *International Conference on Computer Vision*, Santiago, Chile, Dec. 2015.

[26] Y. Wen, Z. Li and Y. Qiao, "Latent factor guided convolutional neural networks for age-invariant face recognition," in *International Conference on Computer Vision and Pattern Recognition*, Las Vegas, NV, USA, Jun. 2016.

[27] X. Shu, J. Tang, Z. Li, H. Lai, L. Zhang and S. Yan, "Personalized age progression with bi-level aging dictionary learning," *IEEE Transactions on Pattern Analysis and Machine Intelligence,* vol. 40, no. 4, pp. 905-917, Apr. 2018.

[28] A. Sepas-Moghaddam, V. Chiesa, P. Correia, F. Pereira and J. Dugelay, "The IST-EURECOM light field face database," in *International Workshop on Biometrics and Forensics*, Coventry, UK, Apr. 2017.

[29] E. Adelson and J. Bergen, "The plenoptic function and the elements of early vision," in *Computation Models of Visual Processing*, Boston, MA, USA, MIT Press, 1991, pp. 3-20.

[30] M. Levoy and P. Hanrahan, "Light field rendering," in *23rd Annual Conference on Computer Graphics and Interactive Techniques*, New York, NY, USA, Aug. 1996.

[31] S. Gortler, R. Grzeszczuk, R. Szeliski and M. Cohen, "The lumigraph," in *Annual Conference on Computer Graphics and Interactive Techniques*, New Orleans, LA, USA, Aug. 1996.

[32] D. Dansereau, "Plenoptic signal processing for robust vision in field robotics," PhD Thesis in Mechatronic Engineering, Queensland University of Technology, Queensland, Australia, Jan. 2014.

[33] B. Wilburn, "High performance imaging using arrys of inexpensive cameras," PhD Thesis in Electrical Engineering, Stanford University, Stanford, CA, USA, Dec. 2004.

[34] ISO/IEC JTC 1/SC 29/WG 1 , "JPEG pleno call for proposals on light field coding," ISO/IEC, Geneva, Switzerland, Jan. 2017.

[35] R. Ng, M. Levoy, M. Bradif, G. Duval, M. Horowitz and P. Hanrahan, "Light field photography with a hand-held plenoptic camera," Tech Report CSTR 2005-02, 2005.

[36] "Lytro website," Lytro Inc, 2016. [Online]. Available: https://www.lytro.com. [Accessed Aug. 2018].

[37] T. Shen, H. Fu and J. Chen, "Facial expression recognition using depth map estimation of light field camera," in *International Conference on Signal Processing, Communications and Computing*, Hong Kong, China, Aug. 2016.

[38] M. Turk and A. Pentland, "Eigenfaces for recognition," *Journal of Cognitive Neuroscience,* vol. 3, no. 1, pp. 71-86, Jan. 1991.

[39] B. Peng, H. Yang, D. Li and Z. Zhang, "An empirical study of face recognition under variations," in *International Conference on Automatic Face & Gesture Recognition*, Xian, China, May 2018.

[40] C. Ding and D. Tao, "A comprehensive survey on pose-invariant face recognition," *ACM Transactions on Intelligent Systems and Technology,* vol. 7, no. 3, pp. 37-79, Apr. 2017.

[41] X. Chai, S. Shan, X. Chen and W. Gao, "Locally linear regression for pose-invariant face recognition," *IEEE Transactions on Image Processing,* vol. 16, no. 7, pp. 1716-1725, Jun. 2007.

[42] I. Masi, F. Chang, J. Choi and S. Harel, "Learning pose-aware models for pose-invariant face recognition in the wild," *IEEE Transactions on Pattern Analysis and Machine Intelligence,* vol. 41, no. 2, pp. 379-393, Feb. 2019.

[43] J. Zhao, Y. Cheng, Y. Xu, L. Xiong and J. Li, "Towards pose invariant face recognition in the wild," in *International Conference on Computer Vision and Pattern Recognition*, Salt Lake City, UT, USA, Jun. 2018.

[44] C. Chan, X. Zou, N. Poh and J. Kittler, "Illumination invariant face recognition: A survey," in *Computer Vision: Concepts, Methodologies, Tools, and Applications*, Hershey, PA, USA, Feb. 2018, pp. 58-79.

[45] W. Zhang, X. Zhao, J. Morvan and L. Chen, "Improving shadow suppression for illumination robust face recognition," *IEEE Transactions on Pattern Analysis and Machine Intelligence,* vol. 41, no. 3, pp. 611-624, Feb. 2018.

[46] O. Gupta, D. Raviv and R. Raskar, "Illumination invariants in deep video expression recognition," *Pattern Recognition,* vol. 76, no. 1, pp. 22-35, Apr. 2018.

[47] Y. Zhang, C. Hu and X. Lu, "IL-GAN: Illumination-invariant representation learning for single sample face recognition," *Journal of Visual Communication and Image Representation,* vol. 59, no. 1, pp. 501-513, Feb. 2019.

[48] N. Dagnes, E. Vezzett, F. Marcolin and S. Tornincasa, "Occlusion detection and restoration techniques for 3D face recognition: a literature review," *Machine Vision and Applications,* vol. 29, no. 1, p. 789–813, Jul. 2018.

[49] X. Wei, C. Li, Z. Lei, D. Yi and S. Li, "Dynamic image-to-class warping for occluded face recognition," *IEEE Transactions on Information Forensics and Security,* vol. 9, no. 12, pp. 2035-2050, Sep. 2014.

[50] J. Yang, L. Luo, J. Qian, Y. Tai, F. Zhang and Y. Xu, "Nuclear norm based matrix regression with applications to face recognition with occlusion and illumination changes," *IEEE Transactions on Pattern Analysis and Machine Intelligence,* vol. 38, no. 1, pp. 156-171, Jan. 2017.

[51] L. Best-Rowden and A. Jain, "Longitudinal study of automatic face recognition," *IEEE Transactions on Pattern Analysis and Machine Intelligence,* vol. 40, no. 1, pp. 148-162, Jan. 2018.

[52] A. Krizhevsky, I. Sutskever and G. Hinton, "Imagenet classification with deep convolutional neural networks," in *International Conference on Neural Information Processing Systems*, Lake Tahoe, NV, USA, Dec. 2012.

[53] X. Wu, R. He, Z. Sun and T. Tan, "A light CNN for deep face representation with noisy labels," arXiv:1511.02683, Apr. 2017.

[54] F. Iandola, S. Han, M. Moskewicz, K. Ashraf, W. Dally and K. Keutzer, "SqueezeNet: AlexNet-level accuracy with 50x fewer parameters and <0.5MB model size," arXiv:1602.07360, Nov. 2016.

[55] C. Szegedy, V. Vanhoucke, S. Ioffe, J. Shlens and Z. Wojna, "Rethinking the inception architecture for computer vision," in *International Conference on Computer Vision and Pattern Recognition*, Las Vegas, NV, USA, Jun. 2016.

[56] M. Mehdipour Ghazi and H. Ekenel, "A comprehensive analysis of deep learning based representation for face recognition," in *International Conference on Computer Vision and Pattern Recognition Workshops*, Las Vegas, NV, USA, Jul. 2016.

[57] K. Grm, V. Struc, A. Artiges, M. Caron and H. Ekenel, "Strengths and weaknesses of deep learning models for face recognition against image degradations," *IET Biometrics,* vol. 7, no. 1, pp. 81-89 , Jan. 2018.

[58] J. Liu, A. Shahroudy, D. Xu, A. Chichung and G. Wang, "Skeleton-based action recognition using spatio-temporal LSTM network with trust gates," *IEEE Transactions on Pattern Analysis and Machine Intelligence,* vol. 40, no. 12, pp. 3007-3021, Dec. 2018.

[59] P. Rodriguez, G. Cucurull, J. Gonzalez, J. Gonfaus, K. Nasrollahi, T. Moeslund and J. Xavier Roca, "Deep pain: Exploiting long short-term memory networks for facial expression classification," *IEEE Transactions on Cybernetics,* vol. 99, no. 1, pp. 1-11, Feb. 2017.

[60] J. Donahue, L. Hendricks, M. Rohrbach, S. Venugopalan, S. Guadarrama, K. Saenko and T. Darrell, "Long-term recurrent convolutional networks for visual recognition and description," *IEEE Transactions on Pattern Analysis and Machine Intelligence,* vol. 39, no. 4, pp. 677-691, Apr. 2017.

[61] D. Dansereau, "Light Field Toolbox V. 0.4," [Online]. Available: http://www.mathworks.com/matlabcentral/fileexchange/49683-light-field-toolbox-v0-4. [Accessed Aug. 2018].

[62] P. Werbos, "Backpropagation through time: What it does and how to do it," *Proceedings of the IEEE,* vol. 78, no. 10, pp. 1550-1560, Oct. 1990.

[63] S. Hochreiter, "The vanishing gradient problem during learning recurrent neural nets and problem solutions," *International Journal of Uncertainty, Fuzziness and Knowledge-Based Systems,* vol. 6, no. 2, pp. 107-116 , Apr. 1998.

[64] A. Zeyer, P. Doetsch, P. Voigtlaender, R. Schlüter and H. Ney, "A comprehensive study of deep bidirectional LSTM RNNS for acoustic modeling in speech recognition," in *International Conference on Acoustics, Speech and Signal Processing*, New Orleans, LA, USA, Jun. 2017.

[65] S. Merity, N. Keskar and R. Socher, "Regularizing and optimizing LSTM language models," arXiv:1708.02182, Aug. 2017.

[66] Y. Gal and Z. Ghahramani, "A theoretically grounded application of dropout in recurrent neural networks," in *International Conference on Neural Information Processing Systems*, Barcelona, Spain, Dec. 2016.

[67] N. Keskar, D. Mudigere, J. Nocedal, M. Smelyanskiy and P. Tang, "On large-batch training for deep learning: Generalization gap and sharp minima," in *International Conference on Learning Representations*, Toulon, France, Apr. 2017.

[68] V. Patel, "The impact of local geometry and batch size on convergence and divergence of stochastic gradient descent," arXiv:1709.04718, Sep. 2017.

[69] O. Déniz, G. Bueno, J. Salido and F. De la Torre, "Face recognition using histograms of oriented gradients," *Pattern Recognition Letters,* vol. 32, no. 12, pp. 1598-1603, Sep. 2011.

[70] G. Zhao, T. Ahonen, J. Matas and M. Pietikainen, "Rotation-invariant image and video description with local binary pattern features.," *IEEE*

*Transactions on Image Processing,* vol. 21, no. 4, pp. 1465-1467, Apr. 2012.

[71] L. Wan, N. Liu, H. Huo and T. Fang, "Face Recognition with Convolutional Neural Networks and subspace learning," in *International Conference on Image, Vision and Computing*, Chengdu, China, Jun, 2017.

[72] M. Wani, F. Bhat, S. Afzal and A. Khan, "Supervised deep learning in face recognition," in *Advances in Deep Learning*, Singapore, Singapore,Springer, Mar. 2019, pp. 95-110.

[73] A. Aissaoui, J. Martinet and C. Djeraba, "DLBP: A novel descriptor for depth image based face recognition," in *International Conference on Image Processing*, Paris, France, Oct. 2014.

[74] H. Lu, K. Plataniotis and A. Venetsanopoulos, "MPCA: Multilinear principal component analysis of tensor objects," *IEEE Transactions on Neural Networks,* vol. 19, no. 1, pp. 18-39, Jan. 2008.

[75] S. Marto, N. Monteiro, J. Barreto and J. Gaspar, "Structure from plenoptic imaging," in *International Conference on Development and Learning and on Epigenetic Robotics*, Lisbon, Portugal, Sep. 2017.

[76] N. Monteiro, S. Marto, J. Barreto and J. Gaspar, "Depth range accuracy for plenoptic cameras," *Computer Vision and Image Understanding,* vol. 168, no. 1, pp. 104-117, Mar. 2018.

[77] H. Jeon, J. Park, G. Choe and G. Park, "Accurate depth map estimation from a lenslet light field camera," in *International Conference on Computer Vision and Pattern Recognition*, Boston, MA, USA, Jun. 2015.

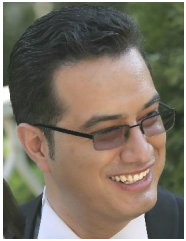

**Alireza Sepas-Moghaddam** (S'11) received the B.Sc degree and the M.Sc. degree with first class honors in Computer Engineering in 2007 and 2010, respectively. From 2011 to 2015, he was with the Shamsipour Technical University, Tehran, Iran, as a lecturer. In 2015, he joined Instituto Superior Técnico (IST), Universidade de Lisboa (UL), Lisboa, Portugal, where he completed the Ph.D. degree with distinction and honour in Electrical and Computer Engineering in 2019. He is now working as a postdoctoral research fellow in the Department of Electrical and Computer Engineering at Queen's University. His main research interests are in the areas of facial image analysis, multi-view feature extraction, and deep learning including convolutional neural, recurrent neural, and generative adversarial networks. He has contributed more than 30 conferences and journal papers.

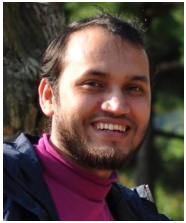

**Mohammad A. Haque** received MS degree in Computer Engineering and Information Technology from Ulsan University, South Korea in 2012 and the PhD degree in Computer Vision from Aalborg University, Denmark in 2016. He is currently with the Visual Analysis of People Lab at the Dept. of Architecture, Design and Media Technology in Aalborg University, Denmark. He was a faculty of the International Islamic University Chittagong, Bangladesh. He started his research career by working on Biometrics and Image Processing. Later he expressed his interest to machine learning and audio-visual signal processing. At present, he is aiming on developing computer vision methods for automatic patient monitoring. He has a long list of publications in the indexed journals and peer-reviewed international conferences. He has also served as a technical committee member in different conferences and reviewer of a number of indexed journals. In the personal life, he is a reader of social cohesion.

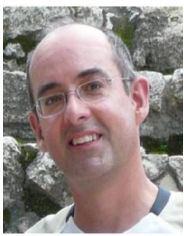

**Paulo Lobato Correia** (M'05–SM'08) obtained the Engineering, MSc and PhD degrees in Electrical and Computer Engineering from Instituto Superior Técnico (IST), Universidade de Lisboa (UL), Portugal, in 1989, 1993 and 2002, respectively. He is Assistant Professor at IST-UL and senior researcher and group leader at Instituto de Telecomunicações. He coordinated the participation in several national and international research projects dealing with image and video analysis and processing. He is Subject Editor (for Multimedia papers) of the Elsevier Signal Processing Journal (since 2018) and Associate Editor of IET Biometrics (since 2013). He was associate editor of the IEEE Transactions on Circuits and Systems for Video Technology (2006-2014) and of the Elsevier Signal Processing Journal (2005-2017). He has been guest editor of several special issues of scientific journals and cooperated in many conference organizing committees, as general chair, program chair, finance chair or special sessions chair. He is a founding member of the Advisory Board of EURASIP, and he is the elected chairman of EURASIP's Special Area Team on Biometrics, Data Forensics and Security. He has co-authored more than 120 conference and journal papers. Research interests include video analysis and processing, with emphasis on biometrical recognition and forensic applications, and intelligent transportation system.

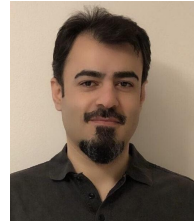

**Kamal Nasrollahi** is currently employed as associate professor at Visual Analysis of People (VAP) Laboratory in Aalborg University where he is involved in several national and international research projects. He obtained his M.Sc. and Ph.D. degrees from Amirkabir University of Technology and Aalborg University, in 2007 and 2010, respectively. His main research interest is on facial analysis systems for which he has published more than 80 peer-reviewed papers on different aspects of such systems in several international conferences and journals. He has won three best conference-paper awards. He has served as technical committee member for several conferences and reviewer for several journals.

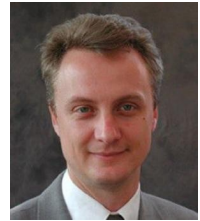

**Thomas B. Moeslund** received his PhD from Aalborg University in 2003 and is currently head of the Visual Analysis of People lab at Aalborg University: www.vap.aau.dk. His research covers all aspects of software systems for automatic analysis of people. He has been involved in 14 national and international research projects, both as coordinator, WP leader and researcher. He has published more than 300 peer reviewed journal and conference papers. Awards include a Most Cited Paper award in 2009, a Best IEEE Paper award in 2010, a Teacher of the Year award in 2010, and a Most Suitable for Commercial Application award in 2012. He serves as associate editor and editorial board member for four international journals. He has co-edited two special journal issues and acted as PC member/reviewer for a number of conferences. Professor Moeslund has co-chaired the following eight international conferences/workshops/tutorials; ARTEMIS'12 (ECCV'12), AMDO'12, Looking at People'12 (CVPR12), Looking at People'11 (ICCV'11), Artemis'11 (ICCV'11), Artemis'10 (MM'10), THEMIS'08 (ICCV'09) and THEMIS'08 (BMVC'08).

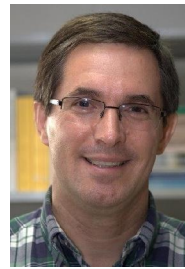

**Fernando Pereira** (S'88–M'90–SM'99–F'08) received the B.S., M.Sc., and Ph.D. degrees in electrical and computer engineering from the Instituto Superior Técnico (IST), Universidade de Lisboa, Lisboa, Portugal, in 1985, 1988, and 1991, respectively. He is currently with the Electrical and Computer Engineering Department, IST, Instituto de Telecomunicações, where he is responsible for the participation of IST in many national and international research projects. He acts often as a Project Evaluator and an Auditor for various organizations. He has authored or coauthored more than 250 papers. His areas of research interest include video analysis, processing, coding and description, and interactive multimedia services. He has been an Associate Editor for the IEEE TRANSACTIONS ON CIRCUITS AND SYSTEMS FORVIDEO TECHNOLOGY, the IEEE TRANSACTIONS ON IMAGE PROCESSING, the IEEE TRANSACTIONS ON MULTIMEDIA, and the IEEE Signal Processing Magazine, and the Editor-in-Chief of the IEEE JOURNAL OF SELECTED TOPICS IN SIGNAL PROCESSING. He is or has been a member of the IEEE Signal Processing Society Image Technical Committees on Video and Multidimensional Signal Processing and Multimedia Signal Processing Technical Committees, and of the IEEE Circuits and Systems Society Technical Committees on Visual Signal Processing and Communications and Multimedia Systems and Applications. He was an IEEE Distinguished Lecturer in 2005. He is an Area Editor of Signal Processing: Image Communication. He has been a member of the Scientific and Program Committees of many international conferences. He has been participating in the work of ISO/MPEG for many years, notably as the Head of the Portuguese delegation, the Chairman of the MPEG Requirements Group, and chairing many ad hoc groups related to the MPEG-4 and MPEG-7 standards.